%% file: 2017_RSS_ProbColl.tex
\documentclass[conference]{IEEEtran}

\input{preamble}

\begin{document}

\title{Uncertainty-Aware Reinforcement Learning for Collision Avoidance}

\author{\authorblockN{Gregory Kahn\authorrefmark{1},
Adam Villaflor\authorrefmark{1},
Vitchyr Pong\authorrefmark{1}, 
Pieter Abbeel\authorrefmark{1}\authorrefmark{2},
Sergey Levine\authorrefmark{1}}
\authorblockA{\authorrefmark{1}Berkeley AI Research (BAIR), University of California, Berkeley}
\authorblockA{\authorrefmark{2}OpenAI}}



%


\maketitle

\begin{abstract} 
Reinforcement learning can enable complex, adaptive behavior to be learned automatically for autonomous robotic platforms. However, practical deployment of reinforcement learning methods must contend with the fact that the training process itself can be unsafe for the robot. In this paper, we consider the specific case of a mobile robot learning to navigate an a priori unknown environment while avoiding collisions. In order to learn collision avoidance, the robot must experience collisions at training time. However, high-speed collisions, even at training time, could damage the robot. A successful learning method must therefore proceed cautiously, experiencing only low-speed collisions until it gains confidence. To this end, we present an uncertainty-aware model-based learning algorithm that estimates the probability of collision together with a statistical estimate of uncertainty. By formulating an uncertainty-dependent cost function, we show that the algorithm naturally chooses to proceed cautiously in unfamiliar environments, and increases the velocity of the robot in settings where it has high confidence. Our predictive model is based on bootstrapped neural networks using dropout, allowing it to process raw sensory inputs from high-bandwidth sensors such as cameras. Our experimental evaluation demonstrates that our method effectively minimizes dangerous collisions at training time in an obstacle avoidance task for a simulated and real-world quadrotor, and a real-world RC car. Videos of the experiments can be found at \url{https://sites.google.com/site/probcoll}.
\end{abstract} 

\IEEEpeerreviewmaketitle


\section{Introduction}
\label{sec:intro}

Policy search via reinforcement learning holds the promise of automating a wide range of decision making and control tasks in safety-critical domains, ranging from self-driving vehicles to drones. However, many reinforcement learning algorithms experience failures at training time, which can be catastrophic in safety-critical domains. Other reinforcement learning algorithms ensure safety by assuming complete state and environment knowledge at training time; however, these assumptions often severely restrict the feasibility of real-world robot deployment. Developing reinforcement learning algorithms that reason about perception and control in unknown environments, understand uncertainty, and explore safely is crucial to deploying reinforcement learning algorithms on safety-critical systems.

One of the central challenges in reinforcement learning is that a robot can only learn the outcome of an action by executing the action itself. Consider a robot learning to navigate an unknown environment while avoiding collisions. This scenario seemingly presents a quandary: the robot needs to learn how to avoid collisions in order to achieve the desired task, but to learn how to avoid collisions, the robot must experience (possibly catastrophic) collisions during training. The robot can overcome this quandary by first experiencing gentle collisions in order to learn about the environment; once the robot is confident about the environment, the robot can avoid catastrophic failures in the future. Central to this approach is that the robot must be able to reason about its own uncertainty because these catastrophic failures are likely to occur in novel scenarios.

Consider an example scenario in which an autonomous drone is learning to fly in an obstacle-rich building. If the drone encounters a novel scenario, the drone will likely crash because the novel scenario is not contained within the training distribution of the reinforcement learning algorithm policy. However, by reasoning about its own policy's uncertainty, the drone can safely interact with the environment and avoid catastrophic failures while also increasing the diversity of its training distribution.

To realize this kind of safe, uncertainty-aware navigation in unknown environments, we propose a model-based learning approach in which the robot learns a collision prediction model and uses estimates of the model's uncertainty to adjust its navigation strategy. By using a speed-dependent collision cost together with uncertainty-aware collision estimates, our navigation strategy naturally chooses to move cautiously when uncertainty is high so as to experience only harmless low-speed collisions, and increases speed only in regions where the confidence of the prediction model is high.

\begin{figure}[t]
\centering
\includegraphics[height=0.128\textheight]{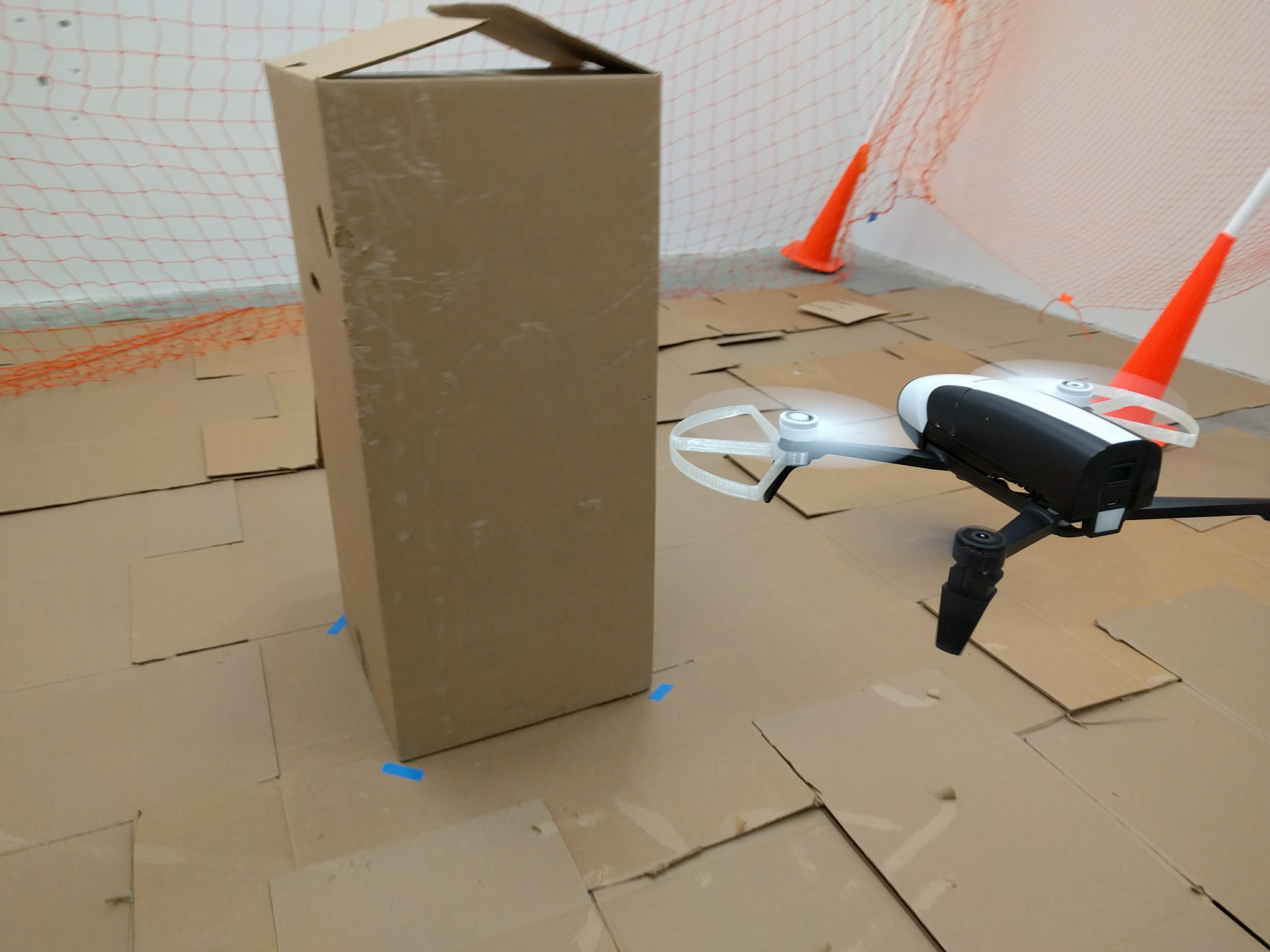}
\includegraphics[height=0.128\textheight]{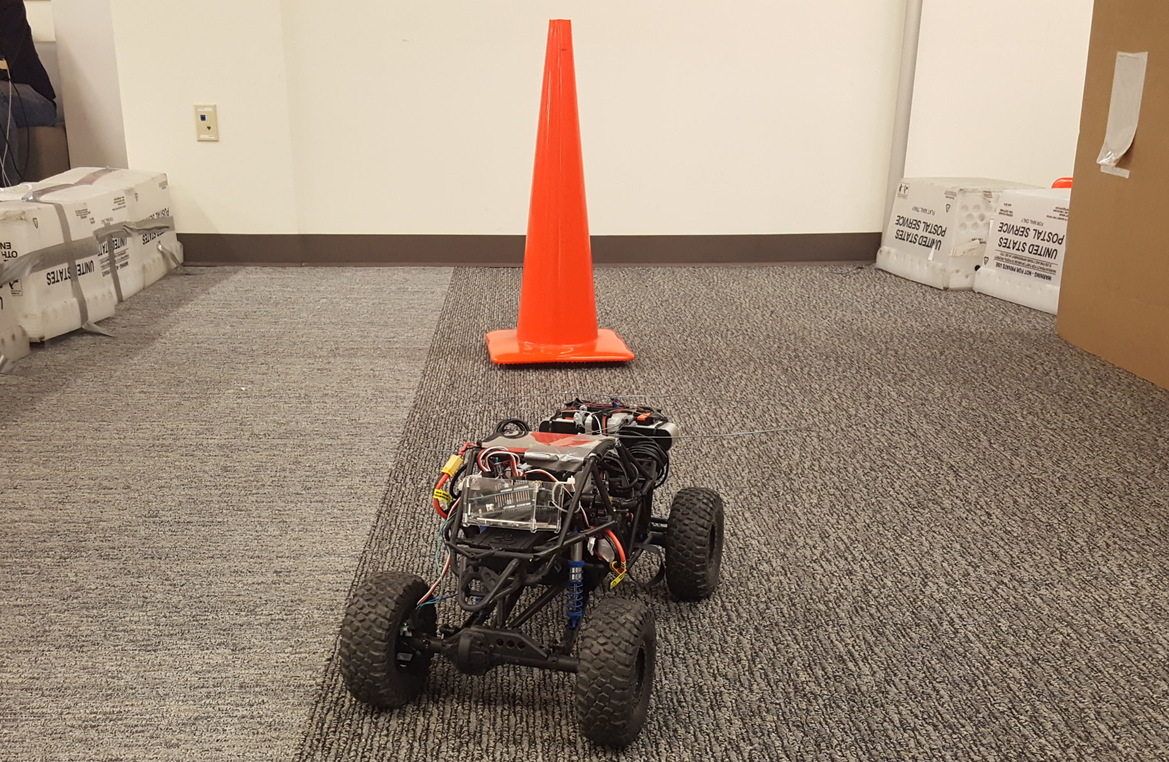}
\caption{\textbf{Uncertainty-aware collision prediction model for collision avoidance}: A quadrotor and an RC car are tasked with navigating in an unknown environment. How should the robots navigate while avoiding collisions? We propose a model-based reinforcement learning approach in which the robot learns a collision prediction model by experiencing collisions at low speed, which is unlikely to damage the vehicle. We formulate a velocity-dependent collision cost that uses collision prediction estimates and their associated uncertainties to enable the robot to only experience safe collisions during training while still approaching the desired task performance.}
\label{fig:teaser}
\vspace*{-20pt}
\end{figure}

Our main contribution is an uncertainty-aware collision prediction model that enables a robot to learn how to accomplish a desired task in an unknown environment while only experiencing gentle collisions. The collision prediction model takes as input the current robot observation and a sequence of controls, computes the probability of a collision occurring along with an estimate of its uncertainty, and outputs a speed-dependent collision cost. The speed-dependent collision cost is a function of the model and its uncertainty, which enables the robot to automatically avoid catastrophic high-speed collisions by acting cautiously in novel situations. We use a deep neural network for the collision prediction model, which allows the model to cope with raw, high-dimensional sensory inputs. To obtain uncertainty estimates from the neural network, we leverage uncertainty estimation methods for discriminatively trained neural networks based on a combination of bootstrapping~\cite{Efron1982_SIAM} and dropout~\cite{Srivastava2014_JMLR,Gal2016_dropout}. A model-based reinforcement learning algorithm then gathers samples using the neural network collision prediction model, which are aggregated and used to further improve the collision prediction model. Our empirical results demonstrate that a robot equipped with our uncertainty-aware neural network collision prediction model experiences substantially fewer dangerous collisions during training while still learning to achieve the desired task. We present an evaluation of our method with various parameter settings for both a simulated and real-world quadrotor, and a real-world RC car (Fig. \ref{fig:teaser}), and demonstrate that our method offers a favorable tradeoff between training-time collisions and final task performance compared to baseline approaches that do not explicitly reason about uncertainty.


\section{Related Work}
\label{sec:related}

In this work, we investigate how model-based reinforcement learning for robot collision avoidance can be made safe and reliable at both training and test time. Reinforcement learning has been applied to a wide range of robotic problems, ranging from locomotion and manipulation to autonomous helicopter flight \cite{kbp-rlrs-13,Deisenroth2011}. Model-free methods have been particularly popular due to their simplicity and favorable computational properties \cite{Peters2006_IROS}. However, model-based methods are generally known to be more sample-efficient \cite{Deisenroth2011_ICML}. In this work, we adopt a model-based approach and learn an uncertainty-aware collision avoidance model; however, similar uncertainty estimation techniques could be extended also to model-free methods.

Several model-based robotic learning algorithms have been proposed that explicitly reason about uncertainty \cite{Deisenroth2011_ICML,Schneider1997_NIPS}. Uncertainty estimates have been used to perform both risk-averse and risk-seeking, optimistic exploration \cite{Moldovan2012_ICML}. The role of uncertainty estimation in our work is to avoid unsafe actions at training time until the model has gained sufficient confidence, which is largely orthogonal and complementary to prior work that seeks to improve exploration in order to accelerate learning. Combining these two directions is a promising direction for future work.

Uncertainty-aware model-based reinforcement learning has been explored in previous work using Bayesian models~\cite{Richter2015_ISRR,Berkenkamp2016_arxiv}. While our work is similar in the overall aim, one of the central goals of our method is to directly process raw inputs from high-bandwidth sensors such as cameras, which necessitates the use of rich and expressive models, such as deep neural networks. Uncertainty estimation for deep neural networks is substantially more challenging, since these models are inherently discriminative. Recent work has proposed to use a Bayesian formulation of neural networks based on dropout~\cite{Gal2016Improving}, as well as to use the bootstrap for exploration \cite{Osband2016_NIPS}, but not, to the best of our knowledge, for uncertainty estimation for the purpose of safety. In this work, we demonstrate that combining both dropout and bootstrap can yield actionable uncertainty estimates for reinforcement learning tasks.

There is much prior work on safe robot control for safety-critical systems such as autonomous cars \cite{Urmson2008_JFR}, legged robots \cite{Wieber2008_IROS}, and quadrotors \cite{Mueller2015_IJRR,Watterson2015_IROS,Gillula2012_ICRA}. A number of recent works have sought to address the question of safety for learning-based robotic systems. Methods based on reachability provide appealing theoretical guarantees, but cannot cope with rich sensory input and are often difficult to scale to high-dimensional systems \cite{Perkins2002_JMLR, Majumdar2016_arxiv, Gillula2012_RSS}. Several works have suggested using discriminative models, including neural networks, to learn safety predictors \cite{Daftry2016_IROS}. These methods generally take the approach of training a model to predict whether an unsafe action will occur, and reverting to a hand-designed safety controller if such a potential failure is detected. Our method offers two advantages over this approach. First, by directly estimating model uncertainty, we do not rely on a discriminative safety estimator. This approach is preferred in environments where the model might encounter previously unseen inputs because a discriminative safety estimator cannot provide meaningful predictions for completely novel inputs; in short, the discriminative safety estimator may erroneously conclude that an unsafe environment is safe. In contrast, a statistical uncertainty prediction such as bootstrapping is more likely to estimate high uncertainty in novel environments. Secondly, our approach does not assume the existence of a manually designed safety control, but instead naturally reverts to more cautious exploratory behavior in the presence of uncertainty. This makes the approach more automated, and does not require a safety mechanism that can recover from arbitrary unsafe situations.

We use deep neural networks to estimate the probability of collision from raw sensory inputs. Combining deep networks with reinforcement learning has been an active area of research in recent years, with applications to video game playing \cite{Mnih2013_NIPS}, control of simulated robots \cite{Schulman2015_ICML,Lillicrap2015_arxiv}, and manipulation \cite{Levine2015_endtoend}. However, most of these applications focus on task complexity or learning speed, rather than explicitly considering uncertainty and safety during training. Prior work has considered safety at training time by using model-predictive control (MPC) with ground truth state information~\cite{Kahn2016_PLATO}. In contrast, our work does not assume any access to ground truth state, which is advantageous for real-world deployment.


\section{Preliminaries}
\label{sec:overview}

Our goal is to control a mobile robot, such as a quadrotor or a car, attempting to navigate an unknown environment. The task may be formally defined in terms of states $\bx$, actions $\bu$, dynamics $\bx_{t+1} = f(\bx_t,\bu_t)$, and observations $\bo$. We use $\mathcal{M}$ to represent the environment, including any potential obstacles. We assume the robot's objective is encoded as a scalar cost function of the form%
\[
\mathcal{C}(\bx_t,\bu_t,\mathcal{M}) = \ctask(\bx_t, \bu_t) + \mathds{1}_{\coll(\bx_t,\mathcal{M})} \ccoll(\bx_t).
\]%
That is, the cost consists of an obstacle-independent task term $\ctask(\bx_t, \bu_t)$, which might include, for instance, flying to a desired position or in a desired direction, as well as an obstacle-dependent collision cost, which is given by the product of an indicator for collision $\mathds{1}_{\coll(\bx_t,\mathcal{M})}$, which is the only term that depends on the environment, and a collision cost $\ccoll(\bx_t)$ that may, for instance, penalize high-speed collisions more than relatively harmless low-speed collisions.

In a fully observable environment where $\mathcal{M}$ is known, the collision indicator can be evaluated exactly, and the problem can be solved by a standard optimal control method, such as the receding-horizon model-predictive control (MPC) approach we use in this work. In receding-horizon MPC, the robot solves an optimal control problem of the form
{\small
\[
\min_{\bu_t,\dots,\bu_{t+H}} \sum_{h=0}^{H} \mathcal{C}(\bx_{t+h},\bu_{t+h},\mathcal{M}) \text{ s.t. } \bx_{t+h+1} = f(\bx_{t+h},\bu_{t+h})
\]
}
at each time step, it executes the action $\bu_t$, advances to time step $t+1$, and repeats the optimization, effectively performing replanning at each time step. In this work, we assume that the dynamics, which might correspond, for instance, to the equations of motion of a quadrotor, are known at least approximately in advance. We instead focus on estimation of the cost, which depends on the unknown environment $\mathcal{M}$. If the environment is unknown and the indicator $\mathds{1}_{\coll(\bx_t,\mathcal{M})}$ cannot be estimated exactly, we can attempt instead to evaluate the probability of a collision using sensor observations, such as LIDAR or camera images. In this case, we can approximate the collision indicator according to
\[
\mathds{1}_{\coll(\bx_{t+h},\mathcal{M})} \approx P(\coll_{t+h} | \bx_t,\bu_{t:t+h} \bo_t).
\]
That is, we can estimate the probability of collision at a future time step $t+h$ based on the current state $\bx_t$, the sequence of actions $\bu_{t:t+h}$ that we intend to take, and the current observation $\bo_t$, which might be used to deduce where the obstacles are located and thereby estimate the probability of collision, without prior knowledge about the environment.

In practice, we will slightly simplify the problem by predicting the probability of a collision at \emph{any} time step $h$ within the MPC horizon $H$. This approximation is not required, but yields a somewhat simpler model that we found performed equally well in practice, especially for relatively short-horizon MPC problems where $\ccoll(\bx_{t+H})$ doesn't change much over the MPC horizon. In this case, the full approximate cost at time $t+h$ evaluated using observation at time step $t$ is given by
\begin{align*}
\mathcal{C}(\bx_{t+h},\bu_{t+h}) \approx\, &\ctask(\bx_{t+h}, \bu_{t+h}) + \label{eqn:main-cost} \\
&\pth(\coll | \bx_{t},\bu_{t:t+H}, \bo_t) \ccoll(\bx_{t+H}),\nonumber
\end{align*}
where we parameterize the probability of collision by model parameters $\theta$, which corresponds to a class of parameteric conditional models. In our case, we present $P_\theta(\coll | \bx_t,\bu_{t:t+H}, \bo_t)$ with a neural network that outputs the parameter of a Bernoulli random variable, as we will discuss in Section~\ref{sec:nn}. Our goal now is to learn the probability of collision model $\pth$ in such a way that avoids catastrophic failures (i.e., high-speed collisions) at both training and test time. However, for the robot to be able to act appropriately in novel situations, the robot must be able to reason about the uncertainty of the collision prediction model $\pth$, as we will discuss in the next section.


\section{Uncertainty-aware Collision Prediction}
\label{sec:approach}

The core component of our approach is an uncertainty-aware collision prediction model $\pth$. Training this collision prediction model from experience presents a dilemma: the robot must first experience collisions in order to learn how to avoid collisions. We formulate a speed-dependent collision cost that uses uncertainty-aware collision estimates, resulting in the robot exploring cautiously when uncertainty is high and moving faster when uncertainty is low. This naturally arising behavior enables the robot to learn about collisions without experiencing catastrophic failures, and subsequently use these safe collision experiences to act more aggressively in the future.

An example application domain and desired application of the uncertainty-aware collision prediction model is the following: consider a quadrotor navigation task in which the objective is to fly fast and avoid collisions in an unknown environment. The quadrotor seeks to learn a collision prediction model that takes as input an image and a sequence of velocity commands and outputs the probability of collision. Initially, the quadrotor flies conservatively because the speed-dependent collision cost favors low-speed actions due to high uncertainty estimates of the collision prediction model. While flying conservatively, the quadrotor experiences safe collisions. These safe collisions, coupled with the associated images, are used to train the collision prediction model; the collision prediction model then learns how to associate images and velocity commands with the likelihood of colliding. As the algorithm continues and the collision prediction model uncertainty becomes low enough, the speed-dependent collision cost will favor high-speed flight.

\subsection{Collision Prediction with Uncertainty}
\label{sec:model}

The collision prediction model $\pth$ takes as input the current state $\bx_t$ and observation $\bo_t$, a sequence of $H$ controls $\bu_{t:t+H}$, and outputs the probability the robot experiences a collision within the horizon. We formulate $\pth$ as a discriminative model using the logistic function $\logistic(y) = 1/(1+\exp(-y))$, so that
\[
\pthfull = \logistic\big(\E[\fthfull]).
\]
Here, $\fthfull$ is a random variable that corresponds to the real-valued output of our stochastic discriminatively trained model, which in our case corresponds to a modified neural network model that can produce uncertainty estimates. In general a variety of alternative models, including stochastic Bayesian models, could be used. Under this model, we can also define a risk-averse collision estimator $\pthsafefull$, given by
{\small
\begin{align}
&\pthsafefull = \nonumber \\
&\logistic\big(\E[\fthfull] + \lambdastd \sqrt{\Var[\fthfull]}\big), \label{eqn:log}
\end{align}}
where $\lambdastd$ is a non-negative user-defined scalar and $\fth$ is scalar-valued function of the current state, observation, and a sequence of controls.

The risk-averse collision prediction model $\pthsafe$ accounts for uncertainty using the variance of the function $\fth$: the larger the variance of $\fth$, the less certain the underlying stochastic model is about the probability of collision. The standard model $\pth$ ignores this uncertainty, while the risk-averse model $\pthsafe$ uses the uncertainty to produce a conservative guess about the collision probability. Note that we use the variance of the sigmoid pre-activation value $\fth$, since sigmoid probabilities are always in the range $[0,1]$. Our goal is to increase the conservative estimate of collision if the model $\fth$ is uncertain (has high variance). However, if we use the sigmoid values, we might systematically underestimate the uncertainty. For example, imagine that the expected value of $\fth$ is a large negative number. Then, even if the variance is very large, the sigmoid expectation will be zero, which means that the sigmoid variance will be low. This is because the tails of the sigmoid flatten any variance in the model, making it invisible in situations where the mean prediction is close to 0 or 1. The hyperparameter $\lambdastd$ allows us to set how conservative the risk-averse model $\pthsafe$ should be, which allows the user to make intuitive tradeoffs between safety and task completion.

\subsection{Velocity-Dependent Collision Cost}

Based on the previously defined risk-averse model, we can now formulate a collision cost that will naturally favor slow, cautious exploration in regions of high uncertainty. The particular cost that we use has the form
\begin{equation}
\ccoll(\bx_t) = \lambdacoll \| \vel_t \|^2,\label{eqn:ccollm-cost}
\end{equation}
where $\vel_{t}$ is the robot velocity at time $t$ and $\lambdacoll$ is a non-negative user-defined scalar that weights the relative importance of $\ccoll$ versus $\ctask$. The full cost is then approximated using the risk-averse collision prediction model, according to
\begin{align}
\mathcal{C}(\bx_{t+H},\bu_{t+H}) \approx\, &\ctask(\bx_{t+H}, \bu_{t+H}) + \label{eqn:main-cost} \\
&\pthsafe(\coll | \bx_t,\bu_{t:t+H}, \bo_t) \ccoll(\bx_{t+H}),\nonumber
\end{align}
With $\pthsafe$ and $\ccoll$ defined, let us now confirm that Eqn. \ref{eqn:main-cost} will naturally favor cautious behavior when the collision prediction model is uncertain, and favor more aggressive behavior when the collision prediction model is confident. If the risk-averse collision prediction probability $\pthsafe$ is large, the robot is encouraged to move slowly in order to minimize $\ccoll$. The collision prediction probability $\pth$ is large when $\E[\fth] + \sqrt{\Var[\fth]}$ is large, which occurs whenever the model predicts a collision (i.e., $\E[\fth] \gg 0$) or when the model is uncertain (i.e., $\Var[\fth] \gg 0$). On the other hand, if the risk-averse collision prediction probability is small, corresponding to a confident no-collision prediction, the robot can focus on minimizing $\ctask$ and move at fast speeds. The collision prediction probability $\pthsafe$ is small when $\E[\fth] + \sqrt{\Var[\fth]}$ is small, which occurs when the model predicts no collision (i.e., $\E[\fth] \ll 0$) and the model is certain (i.e., $\Var[\fth] \approx 0$).

\subsection{Neural Network Collision Prediction Model}
\label{sec:nn}

In order to be able to predict collisions from rich, high-dimensional sensory inputs, such as cameras or LIDAR measurements, we will use deep neural networks to estimate the probability of a collision. In the case of a standard deterministic, discriminatively trained neural network, $\fth$ would represent the pre-activation values in the network at the last layer, while $\pth$ is obtained by applying a sigmoidal nonlinearity to the pre-activations. Such a network can be trained on prior trajectories experienced by the robot simply by slicing all prior data into subsequences of length $H$, and inputting the states $\bx_t$, observations $\bo_t$, and the concatenated sequence of controls $\bu_{t:t+H}$ into the model. The probability of collision labels are binary values recorded by the robot indicating whether a collision occurred, and we can obtain the label for each subsequence simply by checking whether a collision occurred between time steps $t$ and $t+H$. The network can then be trained using standard stochastic gradient descent (SGD) with a cross-entropy loss on the final sigmoid output.

While such a model can provide accurate predictions about collision probability in regions of the environment close to the training data, it is inherently discriminative and deterministic. Such a deterministic model does not provide an estimate of its variance, and therefore is not by itself suitable for risk-averse collision prediction.

\subsection{Estimating Uncertainty with Neural Networks}
\label{sec:nn-uncertainty}

Standard predictive neural network models are trained discriminatively, which means that, even though the network might achieve a high accuracy on samples drawn from the same distribution as the training data, it is very difficult to predict how the network would behave on data drawn from a different distribution. While it is possible to train a neural network model that outputs a mean and a variance as its prediction \cite{Daftry2016_IROS}, this model is not in general guaranteed to output high variances for unfamiliar inputs because the network is by definition trained only on the datapoints that are in the training set. Indeed, such a method for estimating variance is only effective at estimating the inherent noise in the data, and the variance estimates are not a meaningful indication of the model's own uncertainty about its predictions. To produce accurate uncertainty estimates for data that is outside of the training distribution, we must explore techniques that go beyond direct discriminative training. In order to obtain accurate uncertainty estimates from our model, we use two techniques: bootstrapping and dropout.

\textbf{Bootstrapping}: Bootstrapping \cite{Efron1982_SIAM,efron1994introduction} is a simple and effective method of estimating model uncertainty using resampling that can be used with any discriminatively trained model. Given a dataset $\mathcal{D}$, $B$ new datasets $\mathcal{D}^{(b)}$ are sampled with replacement from $\mathcal{D}$ such that $|\mathcal{D}^{(b)}| = |\mathcal{D}|$. Then, instead of training a single model $\mathcal{M}$ on the entire dataset $\mathcal{D}$, $B$ different models $\mathcal{M}^{(b)}$ are trained on the datasets $\mathcal{D}^{(b)}$. The output prediction and uncertainty estimates are the sample mean and standard deviation of the outputs from the population of models.

The intuition behind bootstrapping is that, by generating multiple populations (using sampling with replacement) and training one model per population, the models will agree in high-density areas of the population (i.e., low uncertainty regions) and disagree in low-density areas of the population (i.e., high uncertainty regions). This intuition is backed with theoretical guarantees~\cite{Kleiner2012_ICML}. However, for time- and resource-constrained applications such as robotics, usually only a limited number of bootstraps can be used, which often leads to inaccurate estimates of the model uncertainty.

\begin{algorithm}[t]
\caption{\small{Neural net training with bootstrapping and dropout}}
\label{alg:train-bootstrap-dropout}
\begin{algorithmic}[1]
\STATE \textbf{input}: dataset $\mathcal{D} = \{\bx_t^{(i)}, \bu_{t:t+H}^{(i)}, \bo_t^{(i)}\}$, neural network model $\textnormal{NN}$
\FOR{$b=1$ to $B$}
	\STATE Sample a dataset of subsequences $\mathcal{D}^{(b)}$ from the full dataset $\mathcal{D}$ with replacement
	\STATE Initialize neural network $\textnormal{NN}^{(b)}$ with random weights
	\FOR{number of SGD iterations}
		\STATE Sample datapoint $(\bx_t, \bu_{t:t+H}, \bo_t)$ from $\mathcal{D}^{(b)}$		
		\STATE Sample $\textnormal{NN}^{(b)}_d$ by masking the units in $\textnormal{NN}^{(b)}$ using dropout
		\STATE Run forward pass on $\textnormal{NN}^{(b)}_d$ using $(\bx_t, \bu_{t:t+H}, \bo_t)$
		\STATE Run backward pass on $\textnormal{NN}^{(b)}_d$ to get gradient $g^{(b)}_d$
		\STATE Update model $\textnormal{NN}^{(b)}$ parameters using $g^{(b)}_d$
	\ENDFOR
\ENDFOR
\end{algorithmic}
\end{algorithm}

\textbf{Dropout}: Dropout \cite{Gal2016_dropout} is, by comparison, a computationally cheap method to improve uncertainty estimates. Dropout is commonly used to reduce overfitting in neural networks by randomly dropping units from the neural network during training~\cite{Srivastava2014_JMLR}. Specifically, a given unit with dropout is set to 0 with probability $p$ and left as its original value with probability $1-p$ during training. Dropout prevents units from co-adapting (and thus overfitting) too much because different units are sampled for each forward pass, which effectively samples a new, but related, network during each step of training. Given a neural network $\textnormal{NN}^{(b)}$, dropout in effect constructs a new randomized version of this network $\textnormal{NN}^{(b)}_d$ by sampling independent Bernoulli random variables to act as masks on each neuron.

When dropout is used to reduce overfitting, it is only applied during training in order to force the units in the network to cope with stochastic removal of other units. In order to achieve high accuracy at test time, the dropout regularization is removed and all network weights are scaled by $p$ to compensate for the increased level of activation. However, Gal and Ghahramani \cite{Gal2016_dropout} showed that dropout can be used to obtain uncertainty estimates at test time by calculating the sample mean and standard deviation of multiple stochastic forward passes of the neural network \textit{using dropout}. In this way, dropout can be viewed as an economical approximation to an ensemble method (such as bootstrapping) in which each sampled dropout mask corresponds to a different model. However, dropout underestimates the uncertainy because it acts roughly as a variational lower bound~\cite{Gal2016_dropout}.

\textbf{Neural Networks with Bootstrapping and Dropout}: Alg. \ref{alg:train-bootstrap-dropout} provides an overview of training neural networks with bootstrapping and dropout. From an initial dataset, multiple datasets are resampled with replacement, along with corresponding neural network model instantiations. While performing stochastic gradient descent on each bootstrap, different units are dropped each time a forward pass occurs; the gradient calculated by backpropagation is then used to update that specific bootstrap model's parameters.

At test time, we can evaluate the mean and variance of the ensemble by performing multiple forward passes on each network $\textnormal{NN}^{(b)}$ using multiple instantiations of the dropout process, corresponding to $\textnormal{NN}^{(b)}_d$. The random function $\fthfull$ then corresponds to sampling a network, sampling a dropout process, and evaluating the output. Thus, using neural networks with bootstrapping and dropout, we can estimate $\E[\fthfull]$ and $\Var[\fthfull]$ for use in the risk-averse model $\pthsafefull$.

\subsection{Reinforcement Learning with Risk-Averse Collision Estimation}
\label{sec:rl-alg}

Alg. \ref{alg:rl} provides an overview of how the uncertainty-aware collision prediction model is used in a model-based reinforcement learning algorithm. Each iteration of the algorithm, the cost function $\mathcal{C}$ is formed using the current uncertainty-aware collision prediction model $\pthsafe$. The model predictive controller then samples trajectories using cost $\mathcal{C}$. These sample trajectories are aggregated into a dataset containing all previous sampled trajectories. Then $\pthsafe$ is trained on the dataset according to Alg. \ref{alg:train-bootstrap-dropout} and the next iteration begins.

\begin{algorithm}[t]
\caption{RL with Risk-Averse Collision Estimates}
\label{alg:rl}
\begin{algorithmic}[1]
\STATE Initialize empty dataset $\mathcal{D}$
\STATE Initialize collision prediction model $\pthsafe$
\FOR{iter=1 to max\_iter}
    \STATE Sample trajectories $\{\tau_i\}$ using MPC with cost $\mathcal{C}$
    \STATE Add samples $\{\tau_i\}$ to $\mathcal{D}$
    \STATE Train $\pthsafe$ using $\mathcal{D}$ (Alg. \ref{alg:train-bootstrap-dropout})
\ENDFOR
\end{algorithmic}
\end{algorithm}


\begin{figure*}
\centering
\includegraphics[width=0.95\textwidth]{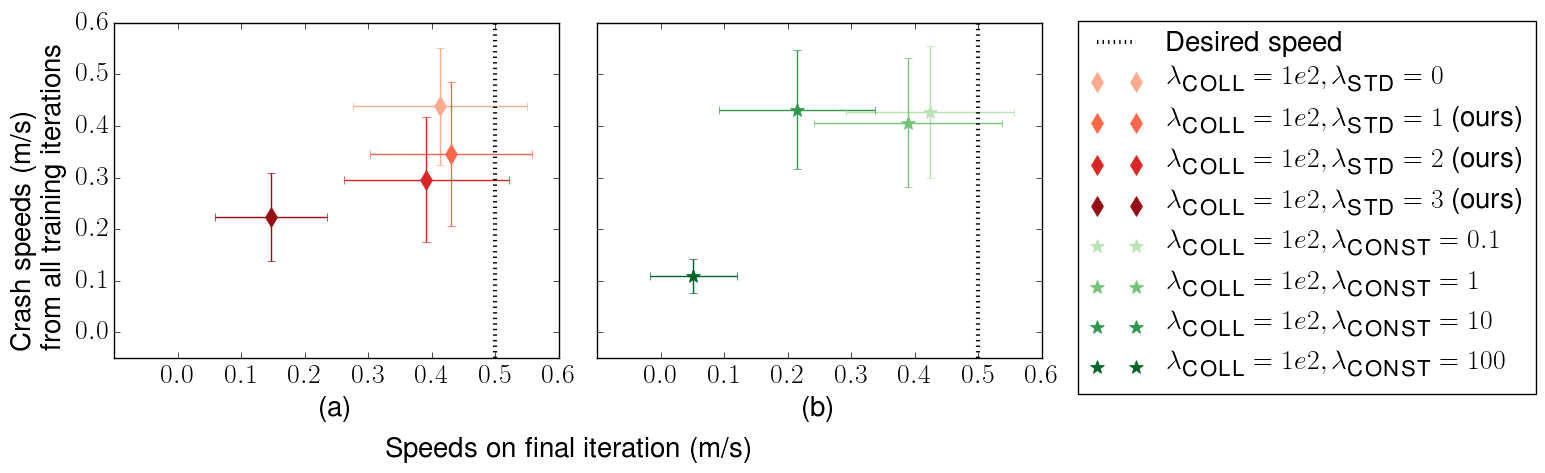}
\caption{\textbf{Comparison of safety versus task performance in simulation}: We investigated the effect of changing parameters in Eqn. \ref{eqn:log} on crash speeds experienced during all training iterations (y-axis) versus the desired objective of flying forward at 0.5 m/s (x-axis). (a) shows the effect of changing $\lambdastd$ for our uncertainty-aware approach. (b) shows the effect of changing $\lambdaconst$ for a conservative baseline, in which the uncertainty in Eqn. \ref{eqn:log} is replaced by a constant. Compared to the conservative baseline approach in (b), (a) shows our uncertainty-aware approach and its parameters can effectively trade off between safety and performance.}
\label{fig:sim-exp}
\vspace*{-15pt}
\end{figure*}

\section{Experiments}
\label{sec:exp}

We present simulated and real-world experiments to evaluate our uncertainty-aware collision prediction model, as well as our proposed model-based RL algorithm. We compare different settings for the parameters in our model, as well as evaluate its performance against a model-based approach that directly estimates the probability of collision, without explicitly accounting for uncertainty. Videos of the experiments can be found at \url{https://sites.google.com/site/probcoll/}.

Our collision prediction model $\pthsafefull$ is a fully connected neural network with two layers with 40 ReLU~\cite{Nair2010_ICML} hidden units each. The activation of the last layer, which outputs the collision probability, is a sigmoid (see Eqn. \ref{eqn:log}). The model inputs are the concatenation of $\bx_t, \bu_{t:t+H}$ and $\bo_t$. We trained the network using ADAM~\cite{Kingma2015_ICLR} and a standard cross-entropy loss. For uncertainty estimation, the simulation experiments used 50 bootstraps and a dropout ratio of $0.2$, while the real-world experiments used 5 bootstraps (due to real-time constraints) and a dropout ratio of $0.05$.

At each time step, the receding-horizon MPC planner chooses among a set of fixed action sequences of horizon length $H$ by evaluating cost $\mathcal{C}$ on each action sequence, and executes the first action of the minimal cost action sequence.

\subsection{Quadrotor experiments}

The simulated and real-world quadrotors have the same states, controls, and observations. We use a high-level representation of the quadrotor in which the control $\bu \in \mathbb{R}^2$ is the commanded planar linear velocity, and therefore we assume the state $\bx$ is estimated such that this level of control is feasible. However, we do not provide the state $\bx$ as input to the collision prediction model. The observation $\bo \in \mathbb{R}^{256}$ is a 16 by 16 grayscale image. The set of action sequences considered by the MPC planner at each time step consists of 190 straight-line, constant-velocity trajectories at various angles and speeds.

\begin{wrapfigure}{r}{0.25\columnwidth}
  \vspace*{-10pt}
  \centering
  \includegraphics[width=\linewidth]{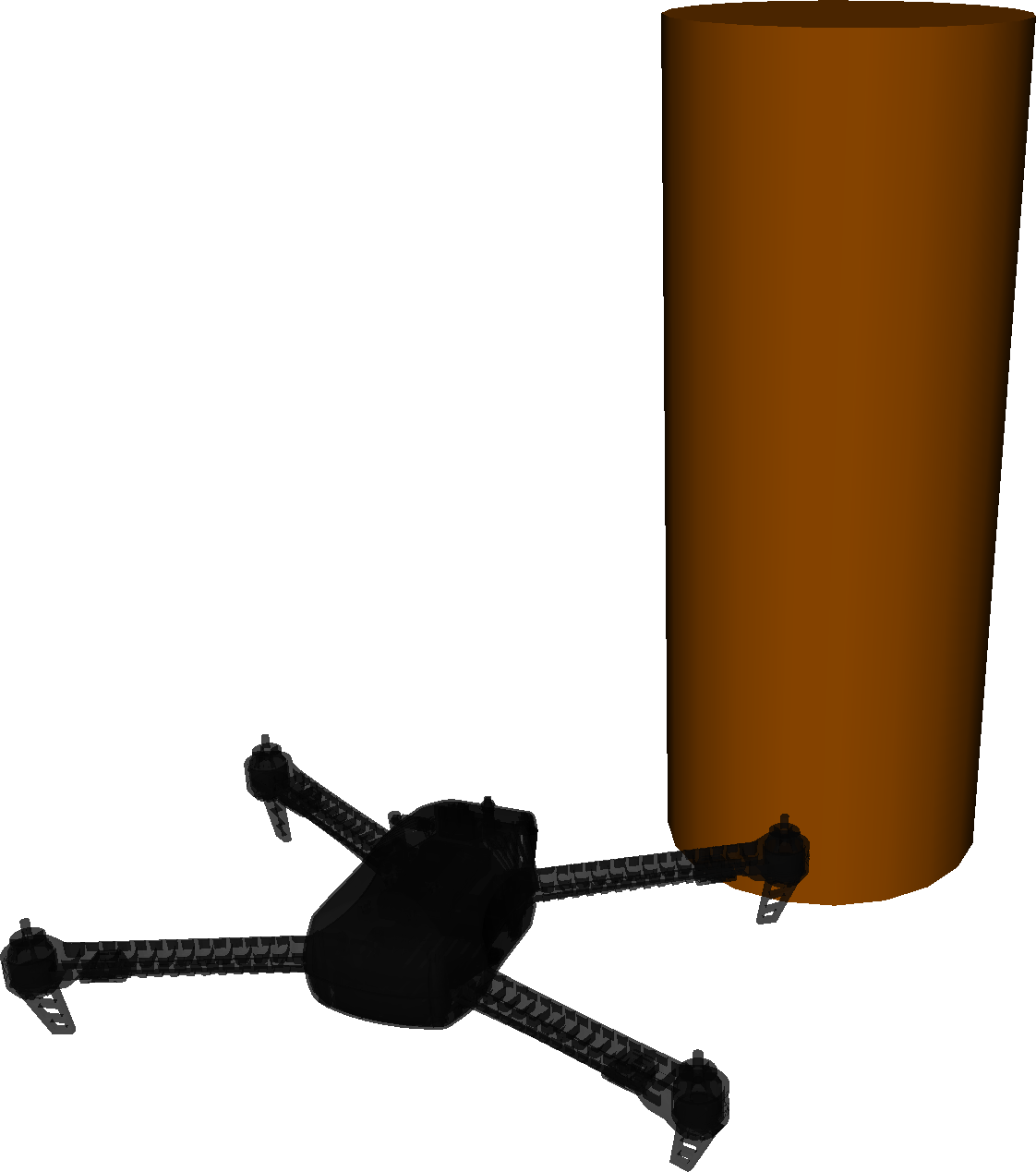}
  \caption{Simulation}
  \label{fig:openrave}
  \vspace*{-15pt}
\end{wrapfigure}
\textbf{Simulated quadrotor}: We first evaluate our uncertainty-aware collision prediction model in a simulated environment consisting of a cylindrical obstacle of radius 0.2m (Fig. \ref{fig:openrave}). The objective $\ctask$ is to fly forward at 0.5 m/s, which is encoded as an $\ell^2$ norm. The time horizon is $H=6$ and each discrete time step corresponds to $\delta = 0.2$ seconds, therefore the planning horizon is $\delta H$ seconds. At each time step, the quadrotor must decide on the sequence of actions using only the observation from a simulated monocular camera.

Fig. \ref{fig:sim-exp} compares safety versus task performance for different variants of Alg. \ref{alg:rl}. All experiments consist of 20 training iterations, with each iteration consisting of 20 on-policy rollouts from start states drawn from the same distribution. Each experiment was run 5 times with different random seeds.

First, we investigate the benefits of incorporating uncertainty into the cost by evaluating different values for $\lambdastd$ (Eqn. \ref{eqn:log}). Fig. \ref{fig:sim-exp}a shows that, when not accounting for uncertainty (i.e., $\lambdastd=0$), the final task performance approaches the desired speed of 0.5 m/s. However, the quadrotor experiences high-speed collisions during training, as shown by the vertical axis. By accounting for uncertainty (i.e., $\lambdastd > 0$), the quadrotor experiences lower speed collisions during training. The final task performance decreases if $\lambdastd$ is increased too much, which is expected: the more conservative the vehicle behaves during training, the longer it takes to learn the task. These results show that $\lambdastd$ allows the user to control their desired degree of risk during training and trade off safety against learning efficiency.

One reasonable question is whether accounting for uncertainty improves safety due to good uncertainty estimates, or simply because adding uncertainty to the collision probability simply makes the vehicle more cautious by penalizing high speeds. To answer this question, we compare our uncertainty-aware approach against a conservative baseline that replaces the uncertainty in Eqn. \ref{eqn:log} with a constant $\lambdaconst$ (Fig. \ref{fig:sim-exp}b). The experiments for $\lambdaconst = 0.1, 1,$ and $10$ show no safety improvement, and also show decreased task performance compared to the baseline $\lambdaconst=0 \equiv \lambdastd=0$. The experiment for $\lambdaconst = 100$ shows substantial safety improvement, but task performance is also substantially diminished. Compared to our uncertainty-aware approach with different settings of $\lambdastd$, the baseline constant penalty approach with $\lambdaconst$ is ineffective at trading off between safety and performance, and always produces overly conservative motions. This indicates that uncertainty estimation is in fact reasoning about the vehicle's surroundings, rather than uniformly encouraging slower flight.

Another reasonable question to ask is whether simply increasing the collision cost $\lambdacoll$ induces safer training behavior. Our experimental results, included in the appendix, show that increasing $\lambdacoll$ does not lead to safer training behavior. Further simulation experiments and results are also provided in the appendix.

\begin{figure*}[t]
\captionsetup[subfigure]{labelformat=empty}
\begin{subfigure}[t]{0.9\textwidth}
	\subcaptionbox{}{\rotatebox[origin=t]{90}{Iteration 0}}
	\includegraphics[height=0.095\textheight]{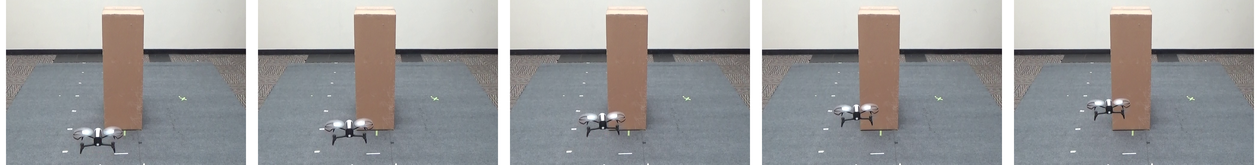}
	\label{fig:bebop-images-itr0}
	\vspace*{-10pt}
\end{subfigure}
\hfill
\begin{subfigure}[t]{\textwidth}
	\subcaptionbox{}{\rotatebox[origin=t]{90}{Iteration 1}}
	\includegraphics[height=0.095\textheight]{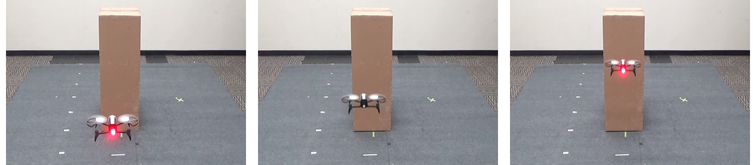}
	\label{fig:bebop-images-itr1}
	\vspace*{-10pt}
\end{subfigure}
\hfill
\begin{subfigure}[t]{\textwidth}
	\subcaptionbox{}{\rotatebox[origin=t]{90}{Iteration 2}}
	\includegraphics[height=0.095\textheight]{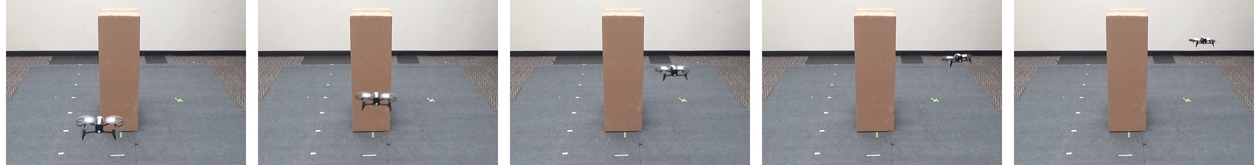}
	\label{fig:bebop-images-itr2}
	\vspace*{-10pt}
\end{subfigure}
\caption{\textbf{Real-world quadrotor experiments}: A Parrot Bebop 2 quadrotor learns to fly while avoiding obstacles using our uncertainty-aware reinforcement learning for collision avoidance algorithm (Alg. \ref{alg:rl}). Sample trajectories from the RL algorithm are shown above. On iteration 0, the quadrotor does not collide with the obstacle, but flies slowly. On iteration 1, the quadrotor flies faster, but collides with the obstacle. On iteration 2, the quadrotor avoids the obstacle while flying at high speed.}
\label{fig:bebop-images}
\vspace*{-15pt}
\end{figure*}

\textbf{Real-world quadrotor}: We evaluated our approach in a real-world environment consisting of a single obstacle, in which the objective is to fly around the obstacle (Fig. \ref{fig:teaser}). Although the task of avoiding a single static obstacle is relatively simple, it is worth noting that the vehicle must perform this task entirely using real-world training data and only monocular images, while minimizing the number of collisions experienced during training. As such, the task is in fact quite challenging.

We ran our experiments using a Parrot Bebop 2 quadrotor. We used the ROS bebop\_autonomy package, which allows the laptop to send linear velocity commands and receive the onboard images in real-time. The quadrotor's objective $\ctask$ is to fly forward at 1.6~m/s, which is encoded as an $\ell^2$ norm. The time horizon $H=3$ and each time step corresponds to $\delta = 0.5$ seconds.

All experiments consist of 5 training iterations, with each iteration consisting of 5 rollouts from 4 different initial positions. This experimental setup can be viewed in the online video. After each rollout, the quadrotor was manually reset to the next initial state.  Note that this reset was solely done for minimizing experimental confounds for the purpose of evaluation, and is not a requirement of our approach. In principle, the vehicle could simply continue flying around the room and collecting data until good performance is achieved. Each experiment was initialized with 6 flight demonstrations provided by a human pilot. These demonstrations were the exact same for all experiments and consisted of 2 crashes and 4 successful flights around the obstacle. To prevent damage to the quadrotor, particularly for the baselines, a human pilot intervened if a crash was imminent; the algorithm therefore treated each intervention as a collision. Each experiment was run 5 times.

Fig. \ref{fig:bebop-images} shows images of our approach during the training process for an example experiment. In the beginning iterations, the quadrotor makes little progress and experiences collisions. As the RL algorithm progresses, the quadrotor is eventually able to fly around the obstacle at high speed.

\begin{figure}[t]
\centering
\includegraphics[width=\columnwidth]{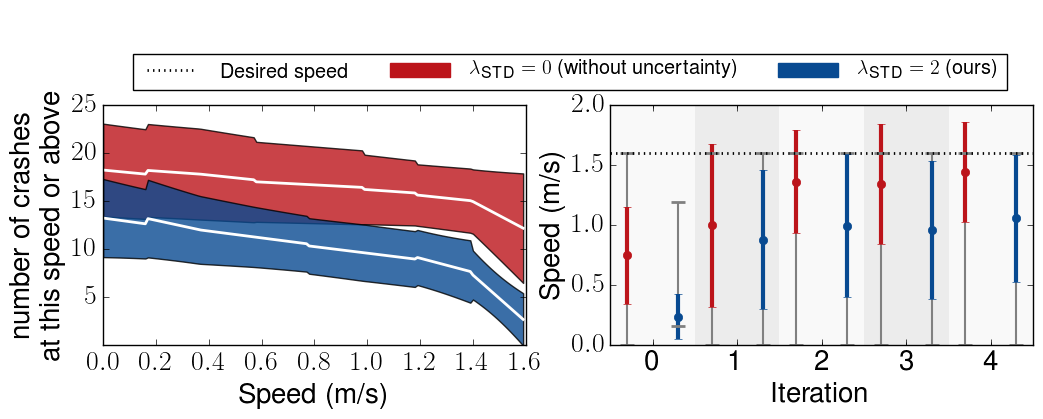}
\caption{\textbf{Comparison of safety versus task performance on a real-world quadrotor}: We investigated the effect of changing $\lambdastd$ in Eqn. \ref{eqn:log} on crash speeds experienced during all training iterations (left) versus the desired objective of flying forward at 1.6 m/s (right) on a Bebop 2 quadrotor. For each value of $\lambdastd$, results are combined from 5 complete experiments with the left plot displaying the mean and std and the right plot displaying the mean, std, and max/min. Increasing $\lambdastd$ leads to fewer crashes (left), but suboptimal performance (right).}
\label{fig:bebop-exp}
\vspace*{-15pt}
\end{figure}

Fig. \ref{fig:bebop-exp} compares safety versus task performance when running our model-based RL algorithm (Alg. \ref{alg:rl}) without uncertainty ($\lambdastd=0$) and with uncertainty ($\lambdastd=2$). When accounting for uncertainty, the quadrotor experiences substantially fewer collisions, especially at higher speeds, but takes longer to approach the desired task performance.

\subsection{Real-world RC car experiments}

\begin{figure*}[t]
\captionsetup[subfigure]{labelformat=empty}
\begin{subfigure}[t]{\textwidth}
	\centering
	\includegraphics[height=0.082\textheight]{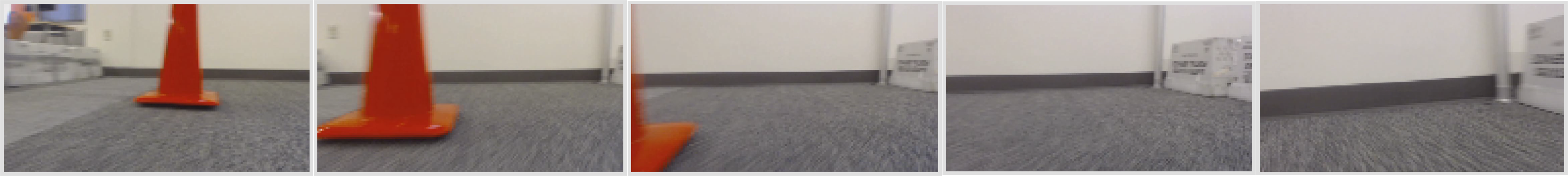}
	\label{fig:rccar-around-montage}
\end{subfigure}
\caption{\textbf{Real-world RC car experiments}: An RC car learns to drive while avoiding obstacles using our uncertainty-aware reinforcement learning for collision avoidance algorithm (Alg. \ref{alg:rl}). A successful rollout is shown above.}
\label{fig:rccar-images}
\end{figure*}

\begin{figure*}[t]
\centering
\includegraphics[width=\textwidth]{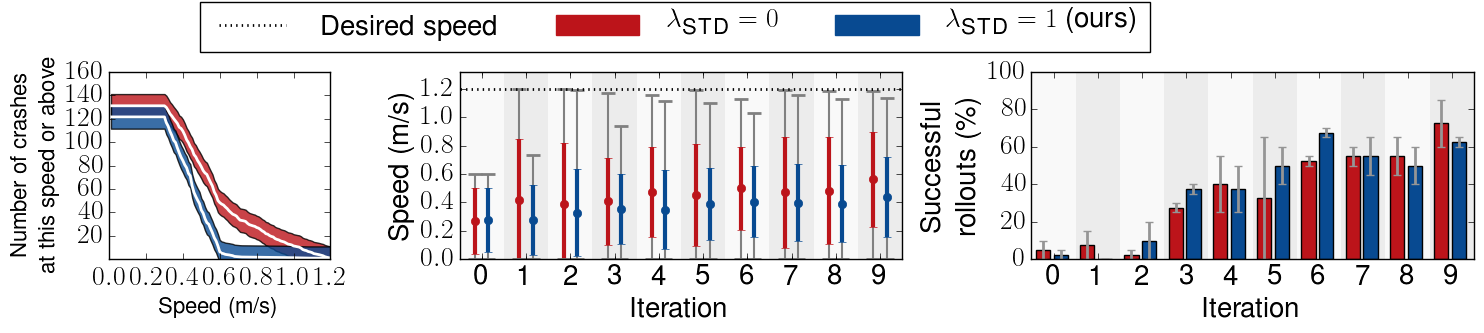}
\caption{\textbf{Comparison of safety versus task performance on a real-world RC car}: We investigated the effect of changing $\lambdastd$ in Eqn. \ref{eqn:log} on crash speeds experienced during all training iterations (left) versus the task objective of driving at 1.2 m/s (middle) and the percentage of rollouts in which the RC car reached the end of the track (right). For each value of $\lambdastd$, results are combined from 2 complete experiments. Our uncertainty-aware approach ($\lambdastd = 1$) experiences 13\% fewer crashes at speeds above 0.6 m/s (left) and comparable task performance (middle/right) compared to the baseline approach which does not account for uncertainty.}
\label{fig:rccar-exp}
\vspace*{-15pt}
\end{figure*}


\begin{wrapfigure}{r}{0.3\columnwidth}
  \vspace*{-30pt}
  \centering
  \includegraphics[width=\linewidth]{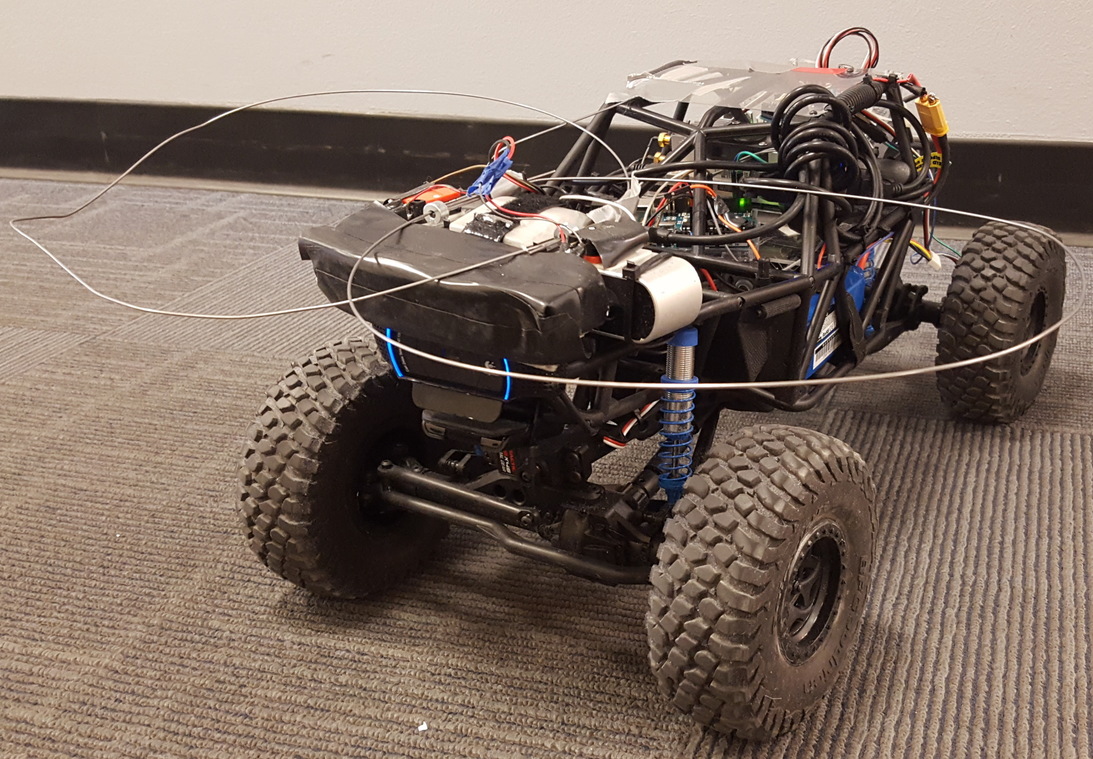}
  \caption{$1/10$th scale RC car with a Logitech C920 Webcam and limit switch collision detectors.}
  \label{fig:rccar-closeup}
  \vspace*{-15pt}
\end{wrapfigure}
We evaluated our approach on an RC car (Fig. \ref{fig:rccar-closeup}) in a simple obstacle avoidance task (Fig. \ref{fig:teaser}). The car is parameterized by control $\bu \in \mathbb{R}^2$ consisting of speed and steering angle and observation $\bo \in \mathbb{R}^{576}$ consisting of a 32 by 18 grayscale image. We do not assume access to any underlying state $\bx$.

The car's objective $\ctask$ is to drive at 1.2~m/s in any direction, which is encoded as an $\ell^2$-norm. The time horizon was set to $H=4$ and each discrete time step corresponds to $\delta = 0.5$ seconds. The set of action sequences considered by the MPC planner at each time step consists of 49 curving, constant-velocity trajectories at various steering angles and speeds.

All experiments consist of 10 training iterations, with each iteration consisting of 5 on-policy rollouts from 4 different initial states. Each rollout ended after either a collision or 10 time steps, therefore each experiment consists of approximately 15 minutes of real-world experience. After each rollout, the car was manually reset to the next initial state. No human demonstrations were used for initialization and each experiment was ran twice. Unlike in the quadrotor experiments, the car was allowed to collide at full speed and automatically registered collisions using limit switches mounted on the front of the car.

Fig. \ref{fig:rccar-images} shows images of our approach during the training process for an example experiment. Initially, the car is unable to avoid the obstacle and side walls, but eventually learns to avoid collisions.

Fig. \ref{fig:rccar-exp} compares safety versus task performance when running our model-based RL algorithm (Alg. \ref{alg:rl}) without uncertainty ($\lambdastd=0$) and with uncertainty ($\lambdastd=1$). The final model-based planner for both approaches succeeds in navigating without colliding for almost 70\% of the rollouts, which is a significant improvement over the initial policy. When accounting for uncertainty, the car experiences fewer high-speed collisions and achieves comparable speeds compared to when not accounting for uncertainty. 


\section{Discussion and Future Work}
\label{sec:future}

We presented a model-based combined perception and control method for learning obstacle avoidance strategies that uses uncertainty estimates to automatically generate safe strategies. Our method is based on predicting the probability of collision conditioned on raw sensory inputs and a sequence of actions, using deep neural networks. This predictor can be used within a model-predictive control pipeline to choose actions that avoid collisions with high probability. In regions of high uncertainty, our risk-averse cost function naturally causes the robot to revert to a cautious low-speed strategy, without any explicit manual engineering of safety controllers or fail-safe mechanisms. We demonstrate our approach is safer compared to methods without uncertainty estimates in both a simulated and real-world quadrotor obstacle avoidance task, as well as a real-world RC car task.

Although our method produces cautious, uncertainty-aware behavior, it does not attempt to explicitly seek out successful strategies except through the MPC optimization. This can cause the algorithm to become stuck in bad local optima. For example, the suboptimal final performance of our approach in the real-world quadrotor experiments with $\lambdastd = 2$ (Fig. \ref{fig:bebop-exp}). A promising direction of future work is to combine our method with optimistic---but still cautious---exploration strategies.

The success of our approach depends strongly on the accuracy of the uncertainty estimates. If the uncertainty estimates are overly optimistic, the robot may experience catastrophic failures. However, if the uncertainty estimates are overly pessimistic, the robot will be perpetually scared and the resulting policy will be suboptimal. This latter case may be another explanation for the suboptimal final performance of our uncertainty-aware approach in the real-world experiments (Fig. \ref{fig:bebop-exp}), therefore future work on developing new uncertainty estimators and characterizing their qualities is important for deploying RL algorithms on robotic systems.

Another promising direction for future work is to generalize our approach beyond collision prediction to other model-based reinforcement learning scenarios. The principle of uncertainty-aware prediction of future events can be readily applied to any feature of the environment, including the expected cost, and exploring this extension to general reinforcement learning problems could produce effective and safe exploration techniques for a wide range of robotic scenarios.


\section{Acknowledgements}

This research was funded in part by the Army Research Office through the MAST program, the National Science Foundation under IIS-1637443 and IIS-1614653, and the Berkeley Deep Drive consortium.



\bibliographystyle{plainnat}
\bibliography{2017_RSS_ProbColl}


\clearpage
\onecolumn
\appendix

\begin{figure}[h!]
\includegraphics[width=\textwidth]{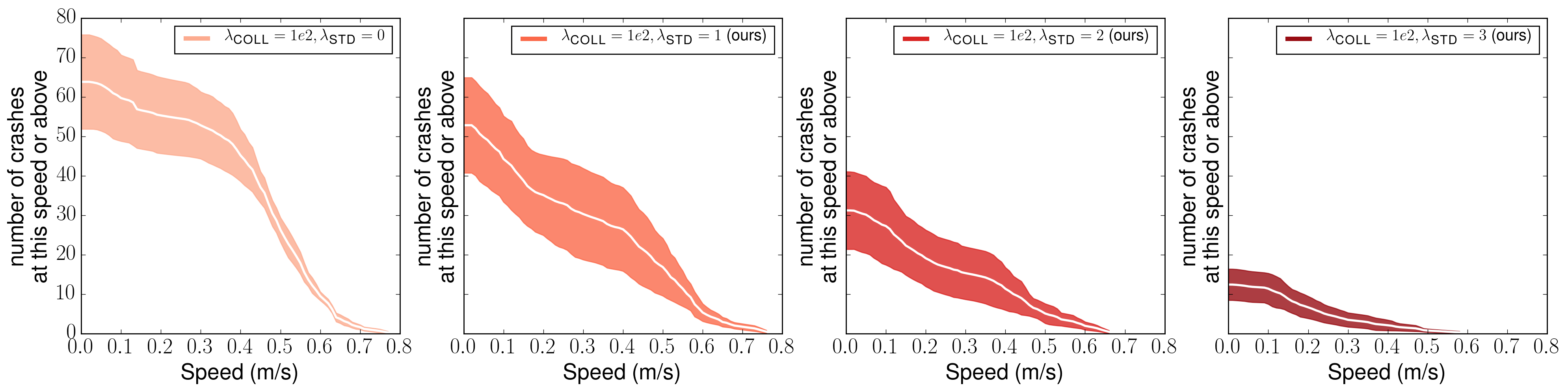}
\includegraphics[width=\textwidth]{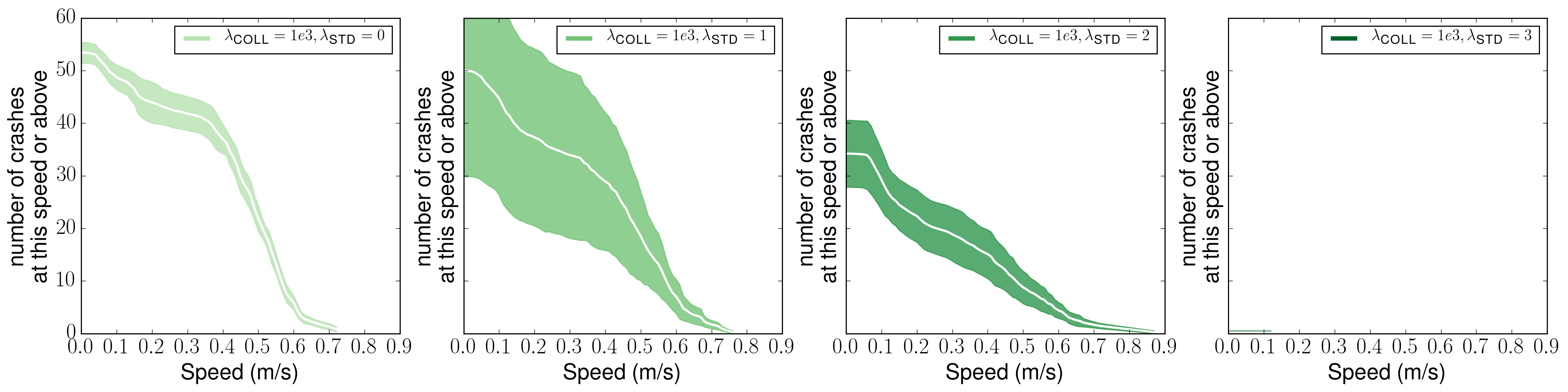}
\includegraphics[width=\textwidth]{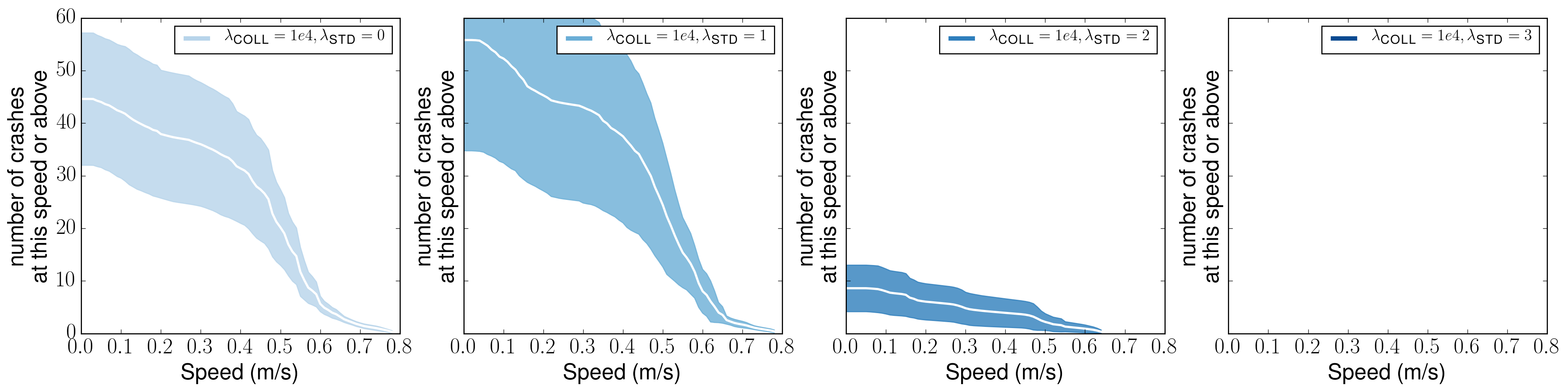}
\includegraphics[width=\textwidth]{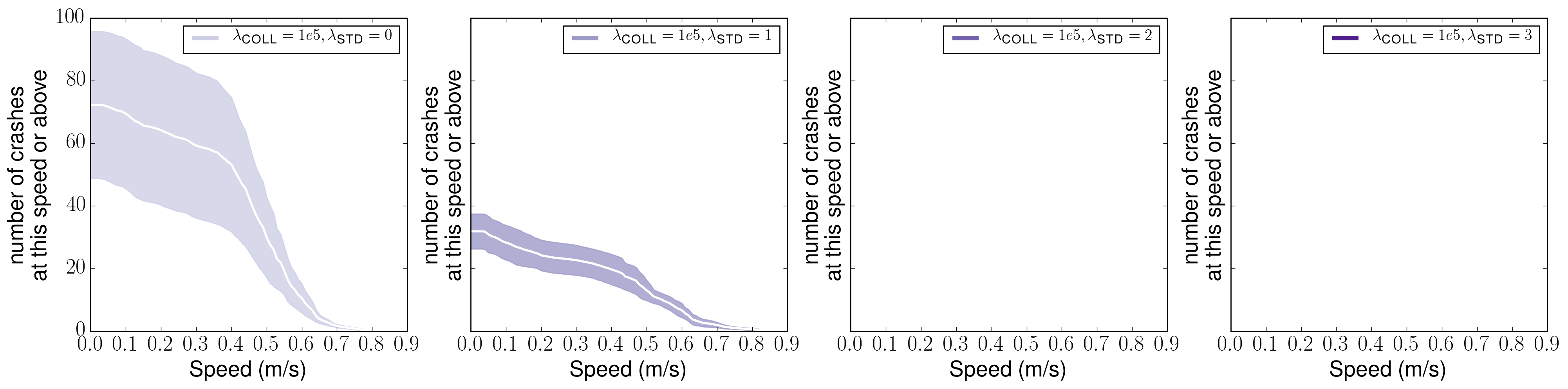}
\caption{\textbf{Safety comparison for different values of $\lambdacoll$ and $\lambdastd$}: Each plot shows the number of training crashes at a given speed or above for a specific setting of $\lambdacoll$ and $\lambdastd$ averaged over 5 experiments. Each row corresponds to a fixed value for $\lambdacoll$ and each column corresponds to a fixed value for $\lambdastd$. Examining the rows show that increasing $\lambdastd$ leads to fewer collisions, and of the collisions that do occur they are at lower speed. Examining the first column shows that increasing $\lambdacoll$ and not accounting for uncertainty does not lead to fewer collisions.}
\label{fig:sim-exp-coll}
\end{figure}

\begin{figure}
\includegraphics[width=\textwidth]{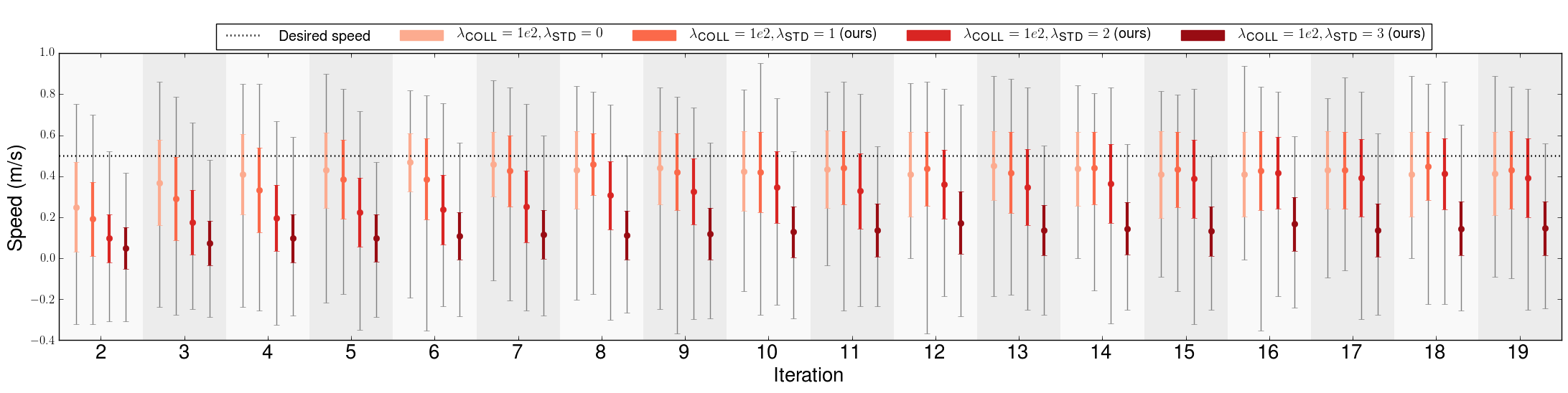}
\includegraphics[width=\textwidth]{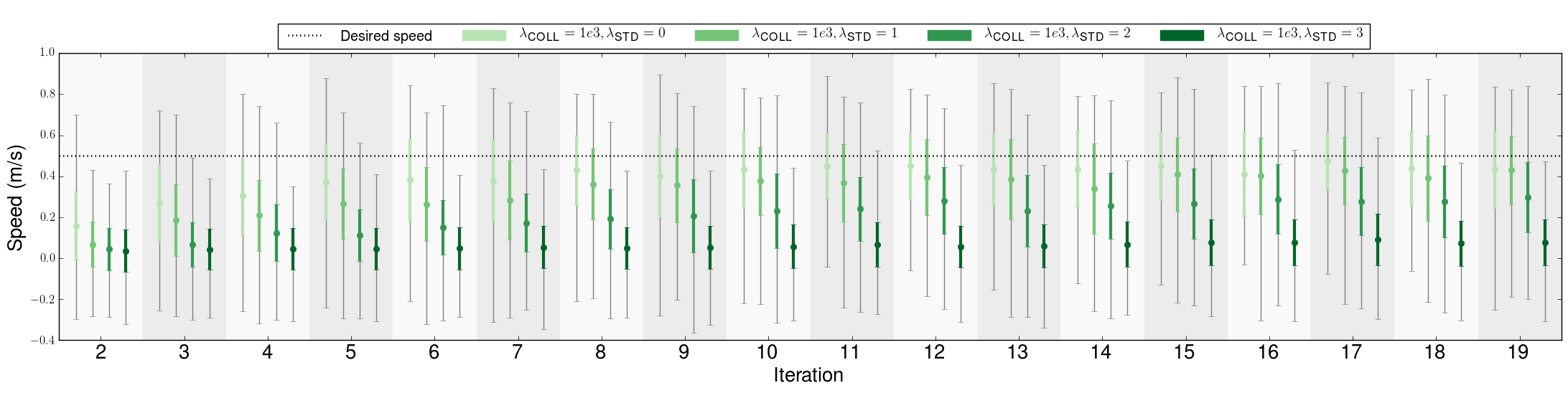}
\includegraphics[width=\textwidth]{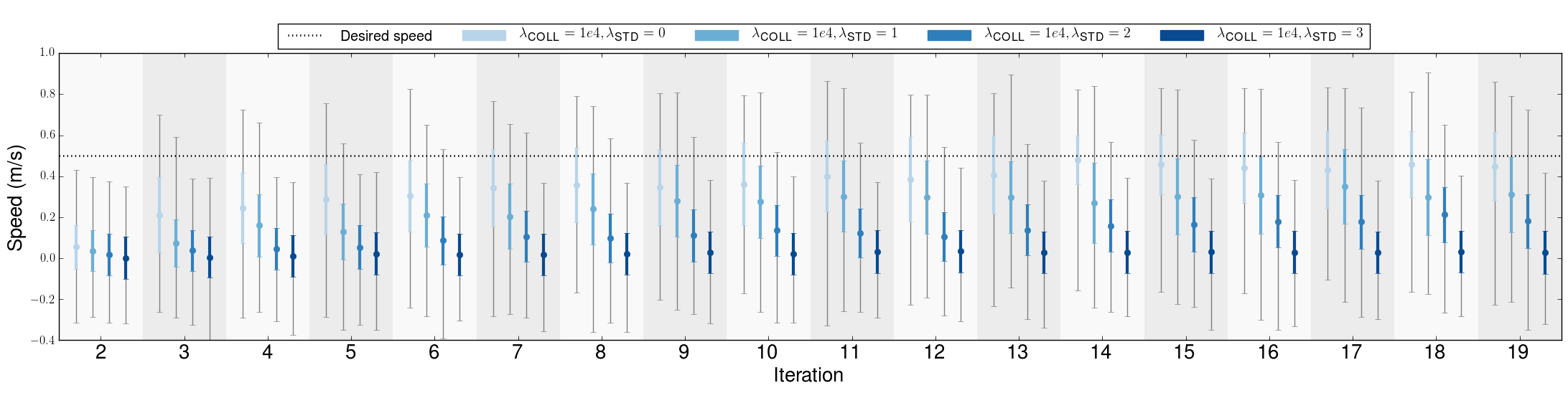}
\includegraphics[width=\textwidth]{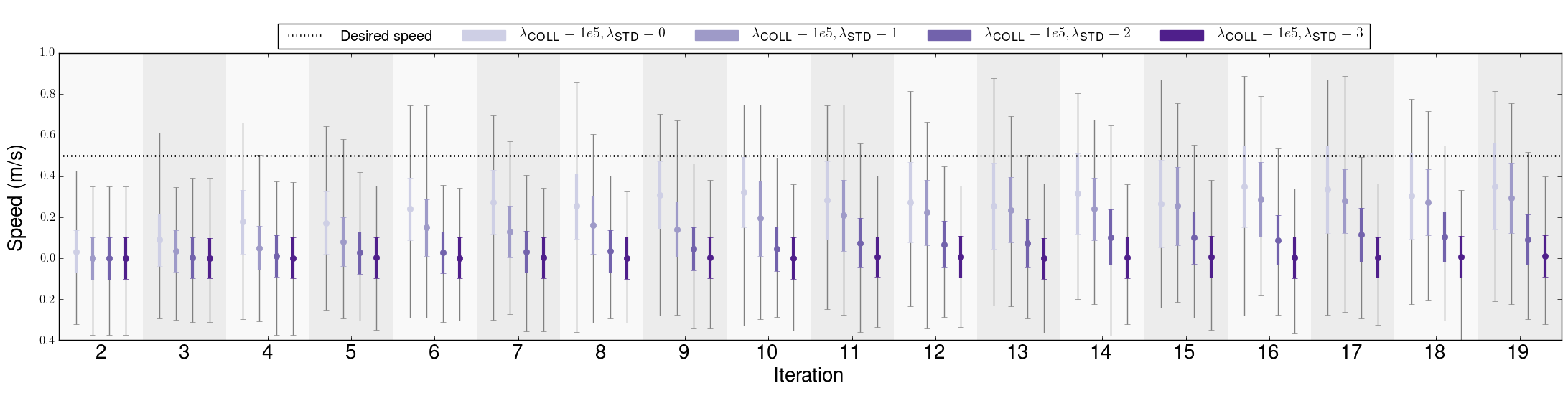}
\includegraphics[width=\textwidth]{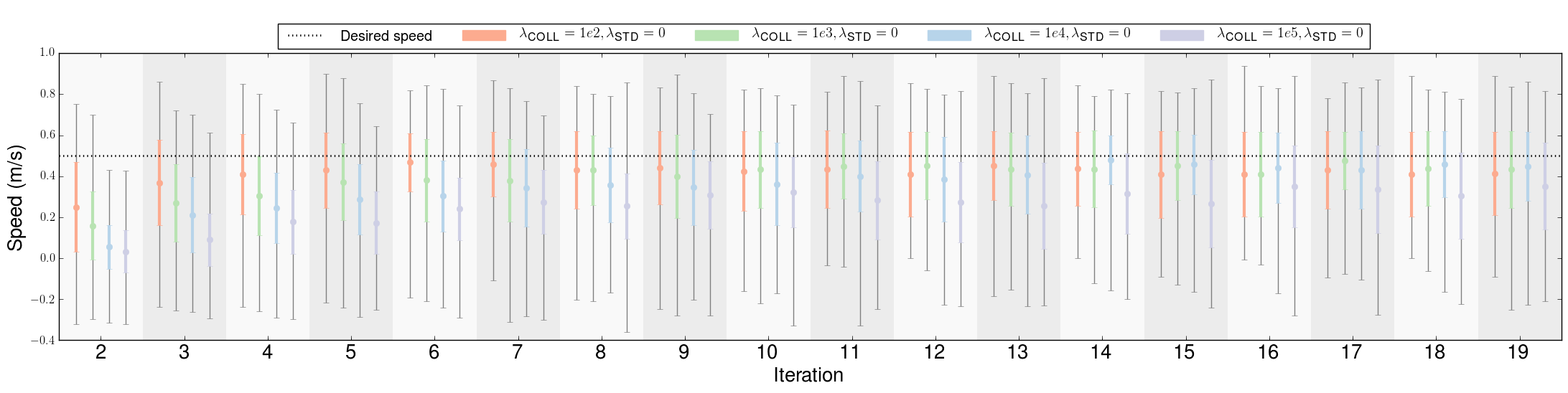}
\caption{\textbf{Comparison of task performance for different values of $\lambdacoll$ and $\lambdastd$}: Each plot shows the task performance at each iteration of the RL algorithm. Results from each setting of $\lambdacoll$ and $\lambdastd$ were averaged from 5 experiments. The top four plots show that higher $\lambdastd$ results in slower progress towards achieving the optimal task performance. Furthermore, for large values of $\lambdastd$, the final performance never reaches the optimal task performance. The bottom plot shows that when not accounting for uncertainty, $\lambdacoll$ has no significant effect on the final task performance.}
\label{fig:sim-exp-xvels}
\end{figure}

\begin{figure}
\includegraphics[width=\textwidth]{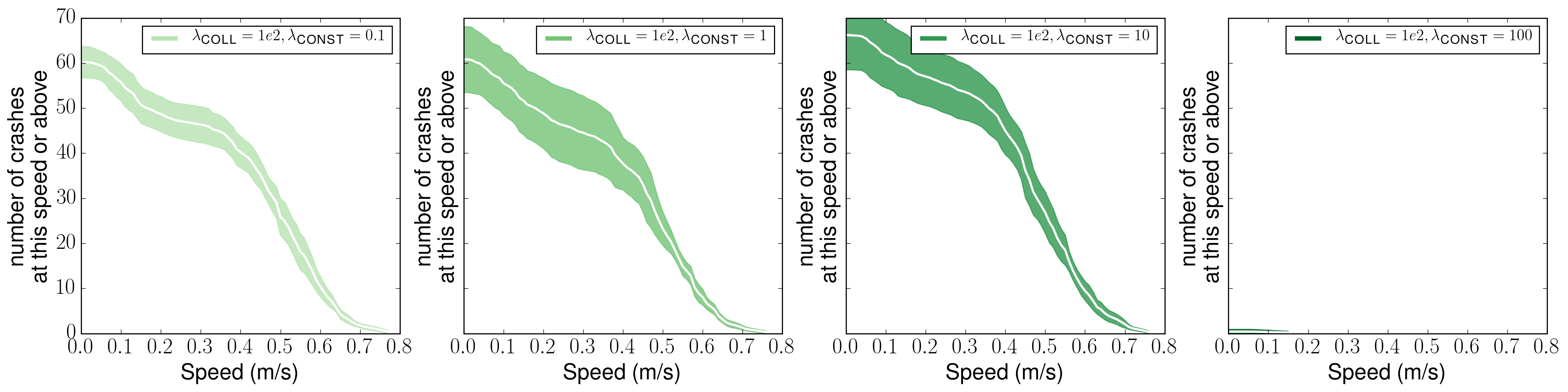}
\includegraphics[width=\textwidth]{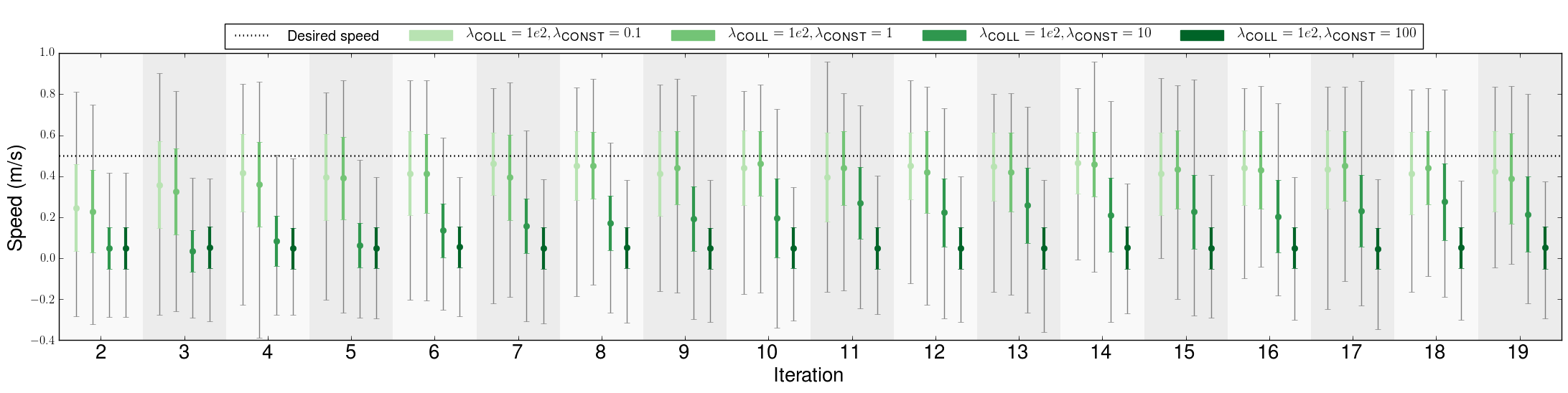}
\caption{\textbf{Comparison of safety versus task performance with a conservative approach}: Further analyzing the data presented in Fig. \ref{fig:sim-exp}, these plots show the effect of changing $\lambdaconst$ for a conservative baseline, in which the uncertainty in Eqn. \ref{eqn:log} is replaced by a constant. The effect of increasing $\lambdaconst$ on the final task performance (bottom) is similar to the effect of increasing $\lambdastd$ in our uncertainty-aware approach, however increasing $\lambdaconst$ does not necessarily increase safety (top).}
\label{fig:sim-exp-coll-xvels-const}
\end{figure}

We present additional results for the simulated quadrotor described in Section~\ref{sec:exp}. We compare the effect of varying the values of $\lambdacoll$ and $\lambdastd$ on safety (Fig. \ref{fig:sim-exp-coll}) and task performance (Fig. \ref{fig:sim-exp-xvels}). Fig. \ref{fig:sim-exp-coll-xvels-const} provides a more detailed analysis of the baseline conservative approach presented in  Fig. \ref{fig:sim-exp}.

\end{document}

%% file: preamble.tex
\usepackage[numbers]{natbib}
\usepackage{multicol}

\usepackage[utf8]{inputenc} 
\usepackage[T1]{fontenc}    
\usepackage[bookmarks=true]{hyperref}       
\usepackage{url}            
\usepackage{booktabs}       
\usepackage{amsfonts}       
\usepackage{nicefrac}       
\usepackage{microtype}      

\usepackage{times}
\usepackage{graphicx} 
\usepackage{caption}
\usepackage{subcaption}
\usepackage{wrapfig}
\usepackage{bbm}
\usepackage{tikz}

\usepackage{algorithm}
\usepackage{algorithmic}

\usepackage{amsmath}
\usepackage{amssymb}
\usepackage{dsfont}

\usepackage[font=footnotesize]{caption,subcaption}
\usepackage{sidecap}

\usepackage{xspace}

\usepackage{color}

\newcommand{\bx}{\mathbf{x}}
\newcommand{\bu}{\mathbf{u}}
\newcommand{\bo}{\mathbf{o}}

\newcommand{\coll}{\textsc{coll}}
\newcommand{\ctask}{\mathcal{C}_{\textsc{task}}}
\newcommand{\ccoll}{\mathcal{C}_{\textsc{coll}}}
\newcommand{\pth}{P_{\theta}}
\newcommand{\pthfull}{\pth(\coll | \bx_t, \bu_{t:t+H}, \bo_t)}
\newcommand{\pthsafe}{\widetilde{P}_{\theta}}
\newcommand{\pthsafefull}{\pthsafe(\coll | \bx_t, \bu_{t:t+H}, \bo_t)}

\newcommand{\lambdastd}{\lambda_{\textsc{std}}}
\newcommand{\lambdaconst}{\lambda_{\textsc{const}}}
\newcommand{\fth}{f_{\theta}}
\newcommand{\fthfull}{\fth(\bx_t, \bu_{t:t+H}, \bo_t)}

\newcommand{\lambdacoll}{\lambda_{\textsc{coll}}}
\newcommand{\vel}{\textsc{vel}}

\newcommand{\E}{\mathds{E}}
\newcommand{\Var}{\mathrm{Var}}
\newcommand{\logistic}{\textsc{L}}

\newcommand{\pushright}[1]{\ifmeasuring@#1\else\omit\hfill$\displaystyle#1$\fi\ignorespaces}

\DeclareGraphicsExtensions{.png}
\graphicspath{ {./figures/} }